\documentclass[conference]{IEEEtran}
\IEEEoverridecommandlockouts
\usepackage{cite}
\usepackage{amsmath,amssymb,amsfonts}
\usepackage{algorithmic}
\usepackage{graphicx}
\usepackage{textcomp}
\usepackage{xcolor}

\usepackage{enumerate}
\usepackage{subcaption}
\usepackage{multirow}
\usepackage{cite}

\begin{document}
%
\title{Open Intent Discovery through Unsupervised Semantic Clustering and Dependency Parsing}


%
\author{\IEEEauthorblockN{Pengfei Liu\IEEEauthorrefmark{1}\thanks{*This paper was published in the 12th IEEE International Conference On Cognitive Infocommunications, 2021. Pengfei Liu is the corresponding author.},
Youzhang Ning\IEEEauthorrefmark{1},
King Keung Wu\IEEEauthorrefmark{1}, 
Kun Li\IEEEauthorrefmark{1} and
Helen Meng\IEEEauthorrefmark{2}}
\IEEEauthorblockA{\IEEEauthorrefmark{1} SpeechX Limited, Shenzhen, China \\
Email: \{pfliu,yzhning,kkwu,kli\}@speechx.cn}
\IEEEauthorblockA{\IEEEauthorrefmark{2} The Chinese University of Hong Kong \& Centre for Perceptual and Interactive Intelligence, Hong Kong SAR, China\\
Email: hmmeng@se.cuhk.edu.hk}
}


\maketitle

\begin{abstract}
Intent understanding plays an important role in dialog systems, and is typically formulated as a supervised learning problem. However, it is challenging and time-consuming to design the intents for a new domain from scratch, which usually requires a lot of manual effort of domain experts. This paper presents an unsupervised two-stage approach to discover intents and generate meaningful intent labels automatically from a collection of unlabeled utterances in a domain.
In the first stage, we aim to generate a set of semantically coherent clusters where the utterances within each cluster convey the same intent. We obtain the utterance representation from various pre-trained sentence embeddings and present a metric of \textit{balanced score} to determine the optimal number of clusters in $K$-means clustering for balanced datasets.
In the second stage, the objective is to generate an intent label automatically for each cluster. We extract the ACTION-OBJECT pair from each utterance using a dependency parser and take the most frequent pair within each cluster, e.g., \textit{book-restaurant}, as the generated intent label.
We empirically show that the proposed unsupervised approach can generate meaningful intent labels automatically and achieve high precision and recall in utterance clustering and intent discovery.
\end{abstract}



%
\IEEEpeerreviewmaketitle

\section{Introduction}

The critical first step towards goal-oriented dialog systems is to determine the goal and the corresponding ontologies, which are typically the natural language understanding targets consisting of intents and slots manually defined by domain experts.
Given the defined ontologies, the tasks of intent recognition and slot filling are usually formulated as supervised learning problems and have achieved high performance in closed-domain datasets, where the training samples are usually hand-labelled.
However, it is challenging and time-consuming to manually define such ontologies accurately and completely when developing real-world dialog systems with potentially a very large volume of conversational data, and almost impossible to know all user intents in advance for complex domains like customer support \cite{perkins2019dialog}.
Therefore, it is desirable to find an automatic method to identify the intents and slots from raw conversational data, in order to reduce the cost in defining the ontologies and annotating the data. This leads to the problem of \textit{open intent discovery}, which has been studied using the methods of supervised learning \cite{cai2017cnn,vedula2019towards}, semi-supervised learning \cite{wang2015mining} and unsupervised learning \cite{perkins2019dialog,chatterjee2020intent}.

This paper investigates the problem of open intent discovery from the perspective of unsupervised learning, i.e., \textit{unsupervised open intent discovery}.
This problem becomes more and more important in developing real-world dialog systems, since it provides a scalable approach to support the scenarios without a predefined taxonomy of intent classes, e.g., a new domain. Besides, it can also be a useful tool for analyzing user intents from social media data such as customer messages, online forums, etc., leading to a great potential of business values \cite{ashkan2009term,wang2013mining}.
The problem of unsupervised open intent discovery involves two tasks: (1) automatically discovering the intents from a corpus of raw conversational data, and (2) automatically assigning each utterance in the corpus to the discovered intent. Such a problem is very challenging as the intents discovered need to be consistent with human understanding, hence each discovered intent should be labelled with an interpretable name. We also observe that it is common in conversational datasets for an utterance to be composed of an ACTION-OBJECT pattern, and thus define an intent as an ACTION-OBJECT pair in an utterance. For example, in the utterance ``\textit{Could you please help me book a flight from Beijing to Shanghai?}", the pair can be defined as (book, flight) which can form a label reflecting the intent, i.e. \textit{book-flight}.

We propose a semantic-based data-driven framework for unsupervised open intent discovery, leveraging recent advances in pre-trained language models \cite{pennington2014glove,peters2018deep,devlin2018bert}. First, we obtain the semantic representation of each utterance using the pre-trained model so that all utterances in the corpus can be represented by points lying in the semantic space. Second, we apply a clustering technique to group the utterances based on their semantic representations by assuming that the utterances of the same intent are close to each other in the semantic space. To better estimate the number of clusters, we introduce a novel metric named \textit{balanced score} which is found to be particularly useful for datasets with intents having similar number of samples. Third, we generate the intent label that are interpretable using the most frequent ACTION-OBJECT pair within each cluster. The ACTION-OBJECT pair for each utterance can be extracted by a dependency parser. We found that the experimental results on the SNIPS dataset are surprisingly good, where the discovered intents match with all the ground-truth intent labels, and the task of assigning each utterance to the appropriate intent achieves a high $F_1$ score of 93\%. 
The contributions of this paper are three-fold:
\begin{enumerate}[(1)]
\item We propose a flexible framework for open intent discovery which can generate human-interpretable intents, and label the utterances with a high $F_1$ in an unsupervised manner; 
\item We develop a rule-based approach for generating intent labels automatically using dependency parsing;
\item We enhance the standard $K$-means clustering in determining the optimal number of clusters for a balanced dataset using a newly proposed \textit{balanced score} metric.
\end{enumerate}

\section{Related Work}

This work on \textit{unsupervised open intent discovery} is closely related with intent recognition, semantic representation and unsupervised clustering. Intent recognition aims to identify the intent expressed in an utterance, e.g., the intent of \textit{buy-computer} in the utterance of ``\texttt{I want to buy a new computer}". This is an important task widely applicable in goal-oriented dialog systems, conversation analysis and online advertisement, where supervised learning methods \cite{tur2011intent,liu2016attention,meng2017dialogue,schuurmans2019intent,khalil2019cross,wang2020dialogue,firdaus2020deep,zhang2016joint} are typically adopted to learn classifiers from labeled intent datasets.
According to different application scenarios, intent recognition can be categorized into (1) query intent classification (e.g., a search engine \cite{jansen2007determining,cao2009context,strohmaier2012acquiring}); (2) intent identification from social media (e.g. Twitter messages) \cite{ashkan2009term,wang2013mining}; (3) user intent understanding in a dialog system \cite{liu2016attention,goo2018slot,niu2019novel}.

We consider semantic representations as continuous vectors in a semantic space which capture the meaning of words, phrases or utterances such that semantically similar words, phrases or utterances are close to each other in the space.
They are typically learned from a large collection of text corpora, e.g., word2vec\cite{mikolov2013distributed}, Glove\cite{pennington2014glove}, Elmo\cite{peters2018deep}, BERT\cite{devlin2018bert}, RoBERTa \cite{liu2019roberta}, Sentence-BERT\cite{reimers2019sentence}, etc.
Unsupervised clustering refers to grouping a collection of unlabeled examples into a number of coherent clusters, where the examples within each cluster are similar based on some similarity metric (e.g., cosine similarity). Typical clustering methods include $K$-means\cite{macqueen1967some}, Gaussian mixture models\cite{reynolds2009gaussian}, DBSCAN\cite{ester1996density}, etc. 
One of the key challenges is to determine the number of clusters, although some have used graphical aids such as the Silhouette score\cite{rousseeuw1987silhouettes}.

Typical approaches to open intent discovery include supervised learning methods, semi-supervised learning methods and unsupervised learning methods.
For example, Vedula et al. \cite{vedula2019towards} proposes a two-stage approach for the task of open intent discovery by first predicting the existence of an intent in an utterance and then tagging utterance with the labels of \textit{Action, Object, and None} using a BiLSTM-CRF model. However, this approach needs a labeled dataset.
Alternatively, Cai et al.\cite{cai2017cnn} adopts a hierarchical clustering method to learn user intent taxonomy, where an intent is represented by a concept pair \textit{(disease, reason)} for a user utterance from online health communities, and proposes a CNN-LSTM attention model to predict user intent.
To address the challenge of the lack of labeled intent datasets, Wang et al.\cite{wang2015mining} introduces a graph-based semi-supervised learning approach to categorize intent tweets into six predefined categories. Unlike the setting in unsupervised learning setting, this approach needs a small number of labeled tweets and a set of intent keywords as input.

From the perspective of unsupervised learning, Perkins and Yang \cite{perkins2019dialog} presents an algorithm named \textit{alternating-view k-means} to discover user intents from query utterances in human-human conversations.
To leverage the whole dialog content, this algorithm divides a dialog into a query view and a content view, which are encoded by two separate neural encoders. The two encoders are updated iteratively using the cluster assignment obtained from the alternative view to encourage them to yield similar cluster assignments for the same user queries.
Along the direction of unsupervised clustering approaches, Chatterjee and Sengupta \cite{chatterjee2020intent} present an intent discovery framework for conversation data, which consists of dialog act classification, density-based clustering, manual annotation of clusters and propagation of intent labels. In comparison with previous approaches, our work differs in three major aspects: (1) we adopt pre-trained language models for utterance representation; (2) we incorporate $K$-means with a penalty term to learn the optimal number of clusters from a balanced dataset; and (3) we generate the cluster labels automatically using a dependency parser.


\section{Problem Formulation}

Given a set of $N$ unlabeled utterances $X =\{x_1, x_2, \ldots, x_N\}$, the objectives of \textit{unsupervised open intent discovery} are: (1) learning a set of $K$ distinct intents and their labels $L =\{l_1, l_2, \ldots, l_K\}$; and (2) generating a collection of labeled utterances $S =\{(x_1, y_1) (x_2, y_2) \ldots, (x_N, y_N)\}$, where $y_n \in L$.
This problem is different from the conventional supervised classification problem since there are no labels in the utterances. Furthermore, there are even no pre-existing intent classes, and thus the problem also differs from the zero-shot intent classification problem \cite{chen2016zero,xia2018zero,liu2019reconstructing}, which needs to know the intent classes in advance although some intents may have no examples in the training dataset. For example, Chen et al. \cite{chen2016zero} proposed an intent expansion framework using a convolutional deep structured semantic model to generate embeddings for both seen and unseen intents, and thus achieved zero-shot intent classification using the expanded unseen intent embeddings.

Following prior literature \cite{vedula2019towards,chen2013identifying,wang2015mining}, we define an intent as an ACTION-OBJECT pair, where ACTION represents a \textit{verb}, \textit{command}, \textit{purpose}, or \textit{a task}, and OBJECT represents \textit{an entity} of word or phrase that the ACTION is going to operate upon.
Different from Vedula et al. \cite{vedula2019towards}, who formulate the extraction of ACTION-OBJECT pairs as a sequence tagging problem that requires labeled tags, we consider this problem from the perspective of unsupervised learning based on clustering and parsing. Although it is difficult to extract the ACTION-OBJECT pairs from certain utterances (e.g, there is no ACTION word in an utterance such as ``A ticket from Paris to London, please!"), we can almost always extract at least one ACTION-OBJECT pair from each cluster. The most frequent ACTION-OBJECT pair or the concatenation of the most frequent ACTION and OBJECT words are used to generate the cluster label. Such a frequency-based method delivers surprisingly good empirical results in automatic label generation (See Section V).

\section{Approach}
\begin{figure*}[htb]
    \centering
    \includegraphics[width=0.72\linewidth]{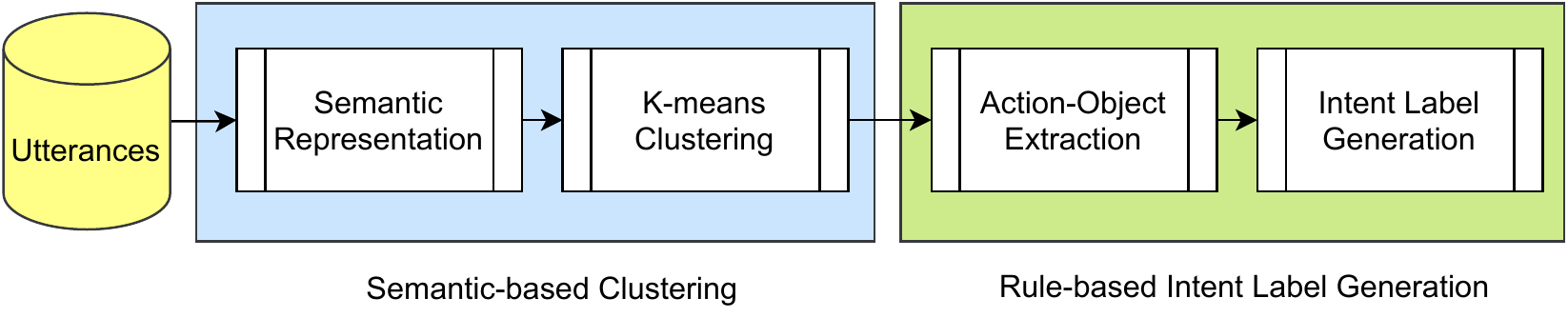}
    \caption{The semantic-based two-stage approach for unsupervised open intent discovery.}
    \label{fig:architecture}
\end{figure*}

We propose a two-stage approach for \textit{unsupervised open intent discovery}: (1) semantic-based unsupervised clustering and (2) parsing-based label generation, as shown in Figure~\ref{fig:architecture}. The first stage consists of deriving semantic representations and performing unsupervised clustering, while the second stage adopts a dependency parser to extract ACTION-OBJECT pairs and generate intent labels in an automatic manner.

\subsection{Semantic Representation}
Pre-trained language models such as Elmo\cite{peters2018deep}, BERT\cite{devlin2018bert}, RoBERTa \cite{liu2019roberta}, Sentence-BERT (SBERT)\cite{reimers2019sentence} have been successfully adopted in various downstream NLP tasks.
Although all these models can be adopted to learn the semantic representation of an utterance, we chose the SBERT model which uses a Siamese network structure to derive sentence embeddings that are particularly suited for efficient semantic similarity search and clustering.
And we evaluate different SBERT representations which are pre-trained from different tasks and datasets, such as natural language inference, semantic textual similarity, paraphrase identification, etc., to find a semantic representation more suitable for the unsupervised learning process.

\subsection{Unsupervised Clustering}
We use the $K$-means algorithm \cite{macqueen1967some} to learn $K$ clusters automatically. For a given utterance represented as a semantic vector $\mathbf {x}$, the objective of $K$-means is to find a set of clusters $C={C_1, C_2, \ldots, C_K}$ which minimizes the within-cluster sum of square errors, as defined in Equation~(\ref{eq:k-means}), where $\boldsymbol{\mu}_i$ is the mean of the examples in the cluster of $C_i$. Note that the $K$-means algorithm tends to generate equal-sized clusters.
\begin{align} \label{eq:k-means}
    \underset {\mathbf {C} }{\operatorname {arg\,min} }\sum _{i=1}^{K}\sum _{\mathbf {x} \in C_{i}}\left\|\mathbf {x} -{\boldsymbol {\mu }}_{i}\right\|^{2}
\end{align}
However, there is a challenge in choosing an optimal $K$ \cite{hamerly2004learning}.
We introduce a metric named \textit{balanced score} to choose the optimal number of clusters. It extends the Silhouette score \cite{rousseeuw1987silhouettes} by a penalty term to encourage a set of clusters with the similar number of examples, as defined in Equations (2-5):
\begin{align}
a(i) &= \frac{1}{|C_i|-1} \sum_{j\in C_i, i\neq j} d(i, j) \\
b(i) &= \min_{k\neq i} \frac{1}{|C_k|} \sum_{j\in C_k} d(i, j) \\
s(i) &= \frac{b(i)-a(i)}{\max\{a(i), b(i)\}} \\
l(i) &= s(i) - \lambda \frac{\sigma}{\mu} \sum_{k=1} ^K \left |\frac{|C_k|}{N} - \frac{1}{K} \right| \label{eq:penalty}
\end{align}

Let $C_i$ denote the assigned cluster of the data point $i$ (i.e., $i \in C_i$), then $a(i)$ measures the mean distance between $i$ and all other data points in the same cluster $C_i$, while $b(i)$ is the smallest mean distance of $i$ to all points in any other cluster. $s(i)$ is defined as the conventional Silhouette score, whereas $l(i)$ extends $s(i)$ by a penalty term to encourage clusters of similar sizes. As defined in Equation~(\ref{eq:penalty}), $\sum_{k=1}^K \left |\frac{|C_k|}{N} - \frac{1}{K} \right|$ becomes 0 if all the clusters have the same number of examples, and becomes larger otherwise, and the ratio of $\frac{\sigma}{\mu}$ is defined as the \textit{coefficient of variation} (cv) \cite{everitt2002cambridge} to measure the dispersion of the cluster sizes having the the standard deviation $\sigma$ and the mean $\mu$. We also use the coefficient $\lambda$, ranging from 0 to 1, to scale the penalty term.  We will empirically show that the penalty term works well with the $K$-means algorithm on a balanced dataset, since both of them prefer the clusters of similar sizes. For imbalanced dataset, the penalty term is not needed and we may adopt a different clustering method such as Gaussian Mixture Models \cite{bouman1997cluster,reynolds2009gaussian}, which have no bias towards learning equal-sized clusters.

\subsection{Label Generation}
Given the $K$ clusters obtained in the previous clustering stage, the most frequent ACTION-OBJECT pair within each cluster is taken as the cluster label, e.g., \texttt{play-music}, \texttt{book-restaurant}.
For each utterance in a cluster, we extract an ACTION-OBJECT pair using the \texttt{dobj} rule parsed from the spaCy dependency parser\cite{honnibal2015improved}, where the ACTION word is a verb and the OBJECT word is a noun or a noun phrase. If there is no such pair in an incomplete or overly short utterance, either ACTION or OBJECT is denoted as NONE. Since we generate the cluster label using the most frequent ACTION-OBJECT pair, the final label for a cluster will not be affected by these incomplete pairs. 
Besides, we filter out the OBJECT words which are numbers to extract the words with clear semantics.
In the ideal cases, the most frequent ACTION-OBJECT pair within each cluster serves as a good label for the cluster, e.g., \texttt{book-restaurant}. In extreme cases, if there is not even a single ACTION-OBJECT pair available, we take the most frequent ACTION word and the most frequent OBJECT word and concatenate them as the cluster label.

\section{Experiments}

\subsection{Experimental Setup}
To evaluate the proposed framework, we use the publicly available SNIPS\cite{coucke2018snips} dataset which has 13,784 sentences and 7 ground-truth intent labels, such as \textit{PlayMusic}, \textit{RateBook}, etc. 
The SNIPS dataset is a good candidate to evaluate the proposed approach, since the manual intent labels generally follow an ACTION-OBJECT pattern and all the intents have similar numbers of utterances (cv=0.26). Note that these intent labels are only used for performance evaluation in our experiments.

For unsupervised clustering, we obtain the semantic representation of each sentence from 4 different models, as shown in Table~\ref{tab:representations}. The first three models are pre-trained using the SBERT model \cite{reimers2019sentence} with various pre-trained tasks using different datasets, while the fourth is called Universal Sentence Encoder \cite{cer2018universal}, which is pre-trained from a variety of web sources such as Wikipedia, web news, discussion forums and achieves superior performance on the semantic textual similarity task.
\begin{table}[htb]
\centering
\caption{Pre-trained sentence representation models.}
\label{tab:representations}
\resizebox{\linewidth}{!}{%
\begin{tabular}{r|ll}
\hline
\textbf{Name} & \textbf{Pre-trained Model} & \textbf{Training Objective} \\ \hline
nli-bert & nli-bert-base-max-pooling & natural language inference \\
stsb-bert & stsb-roberta-base & semantic textual similarity \\
paraphrase & paraphrase-distilroberta-base-v1 & paraphrase identification \\
universal & universal sentence encoder & unsupervised \& NLI \\ \hline
\end{tabular}%
}
\end{table}

\subsection{Evaluation Metrics}
Since the SNIPS dataset has ground-truth intent labels, we adopt precision (P), recall (R) and $F_1$ to evaluate the proposed intent discovery framework.
Following \cite{chatterjee2020intent}, we also use the Normalized Mutual Information (NMI \cite{knops2006normalized}) and Adjusted Rand Index (ARI \cite{rand1971objective}) as the metrics for evaluating clustering performance.
NMI normalizes the mutual information between a predicted clustering and the true clustering, and ranges from 0 to 1; while ARI computes the similarity between two clusterings by considering all pairs of examples and counting the proportion of pairs that are assigned to the same or different clusters.

\subsection{Results and Analysis}

\subsubsection{Stage I: Unsupervised Clustering}
\begin{figure}[htb]
\centering
\begin{subfigure}{0.236\textwidth}
  \centering
  \includegraphics[width=1.1\linewidth]{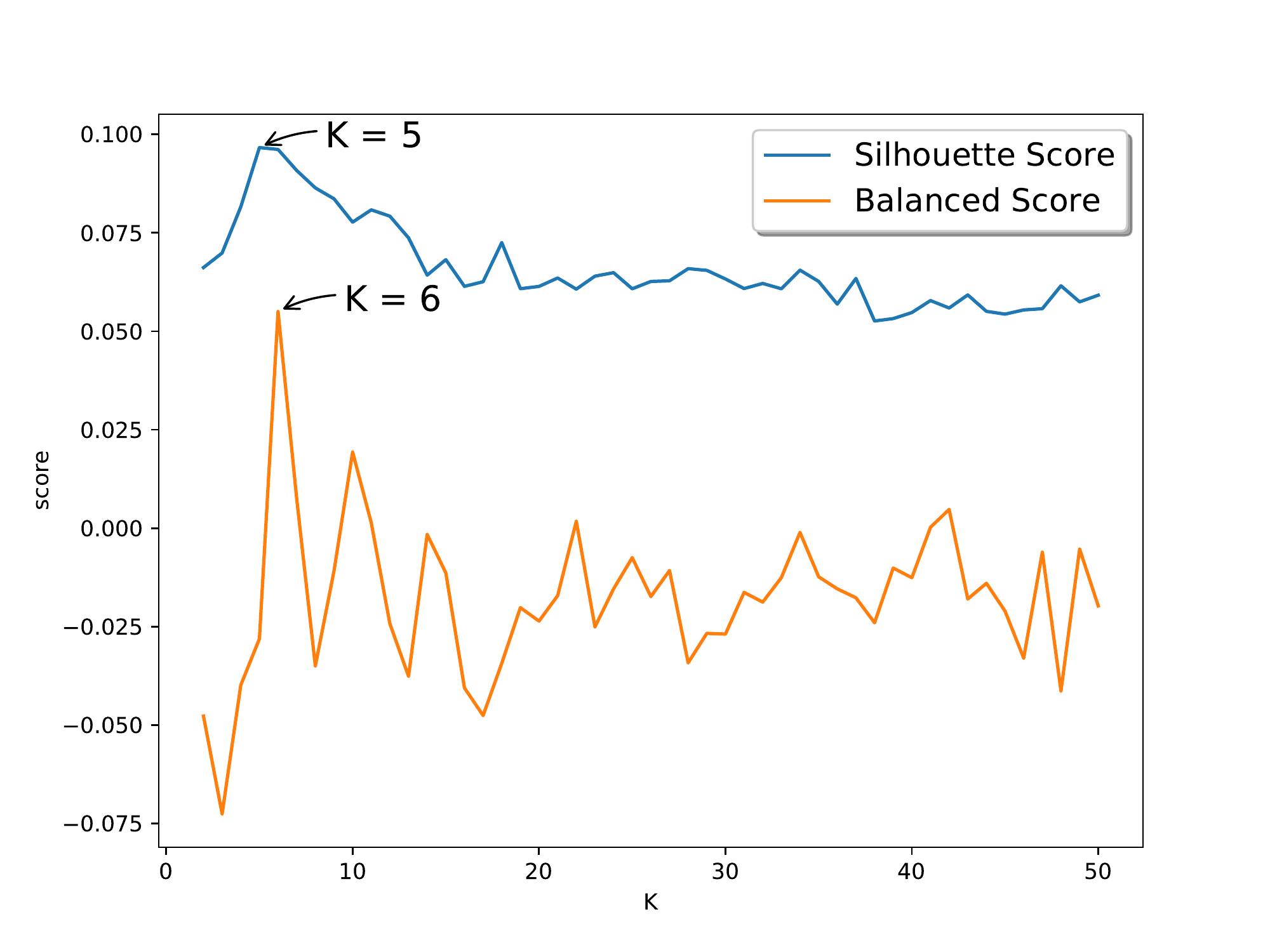}
  \caption{nli-bert}
  \label{fig:nli}
\end{subfigure}
\begin{subfigure}{0.236\textwidth}
  \centering
  \includegraphics[width=1.1\linewidth]{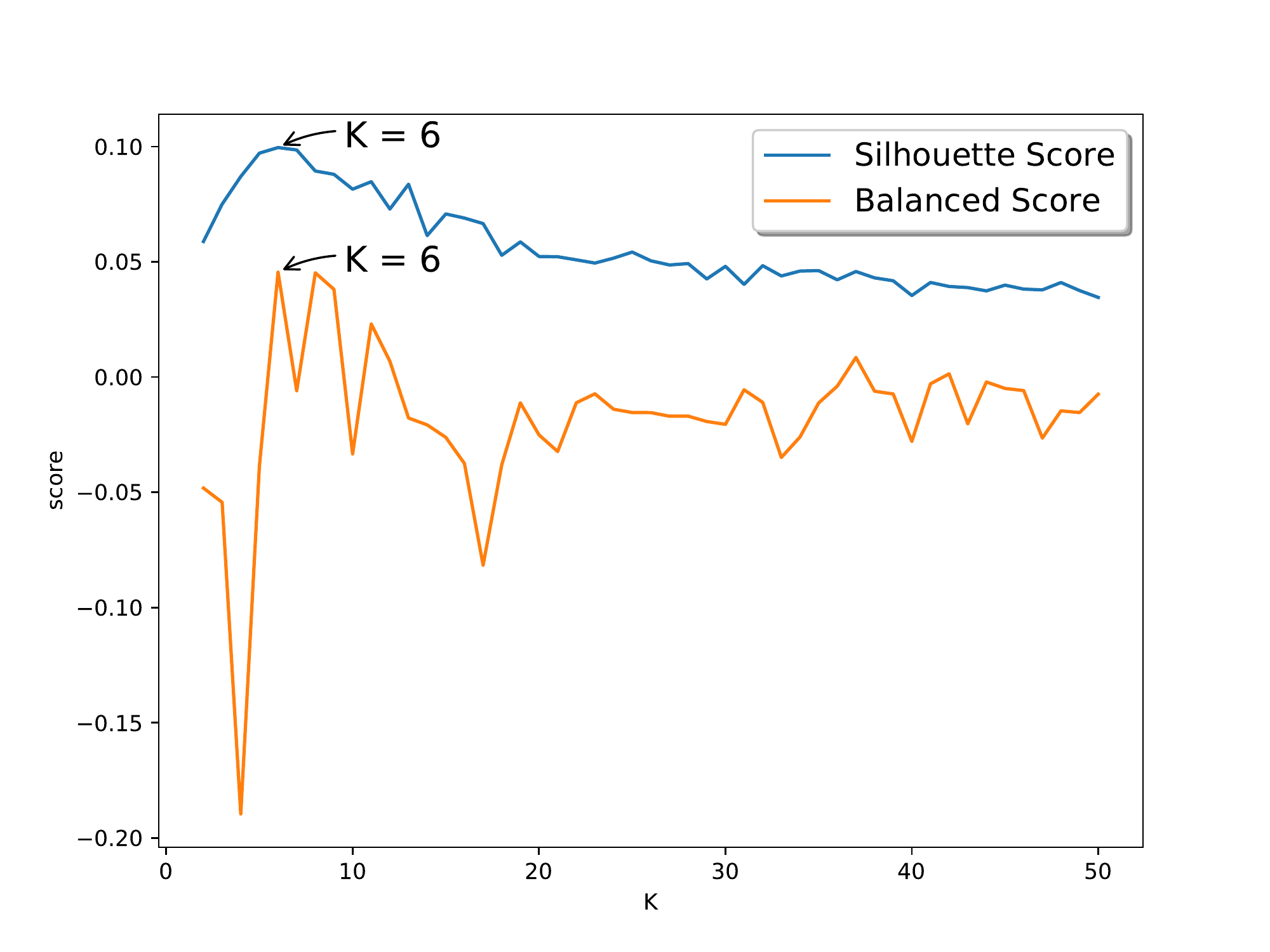}
  \caption{stsb-bert}
  \label{fig:stsb}
\end{subfigure}%

\begin{subfigure}{0.236\textwidth}
  \centering
  \includegraphics[width=1.1\linewidth]{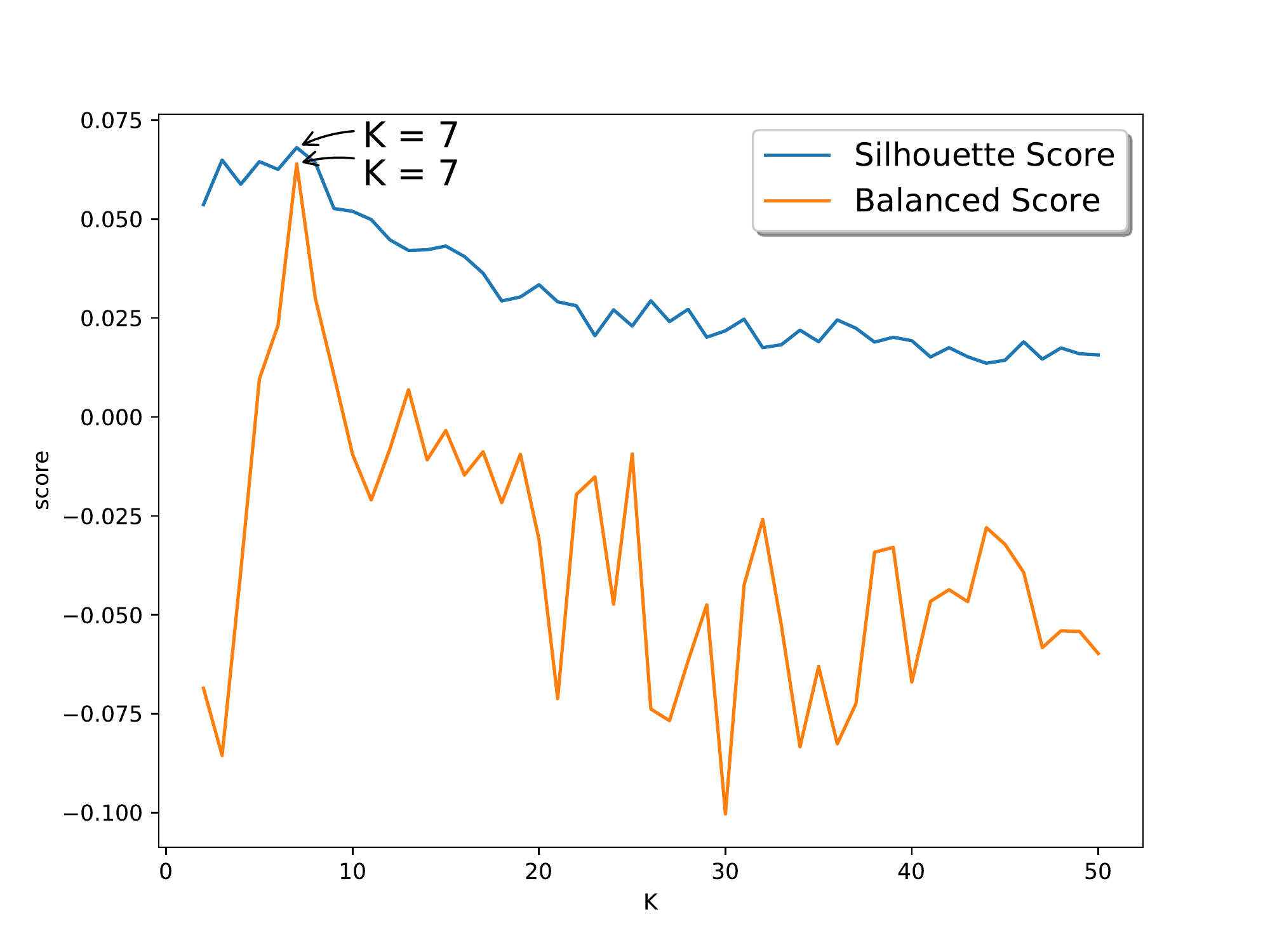}
  \caption{paraphrase}
  \label{fig:paraphrase}
\end{subfigure}
\begin{subfigure}{0.236\textwidth}
  \centering
  \includegraphics[width=1.1\linewidth]{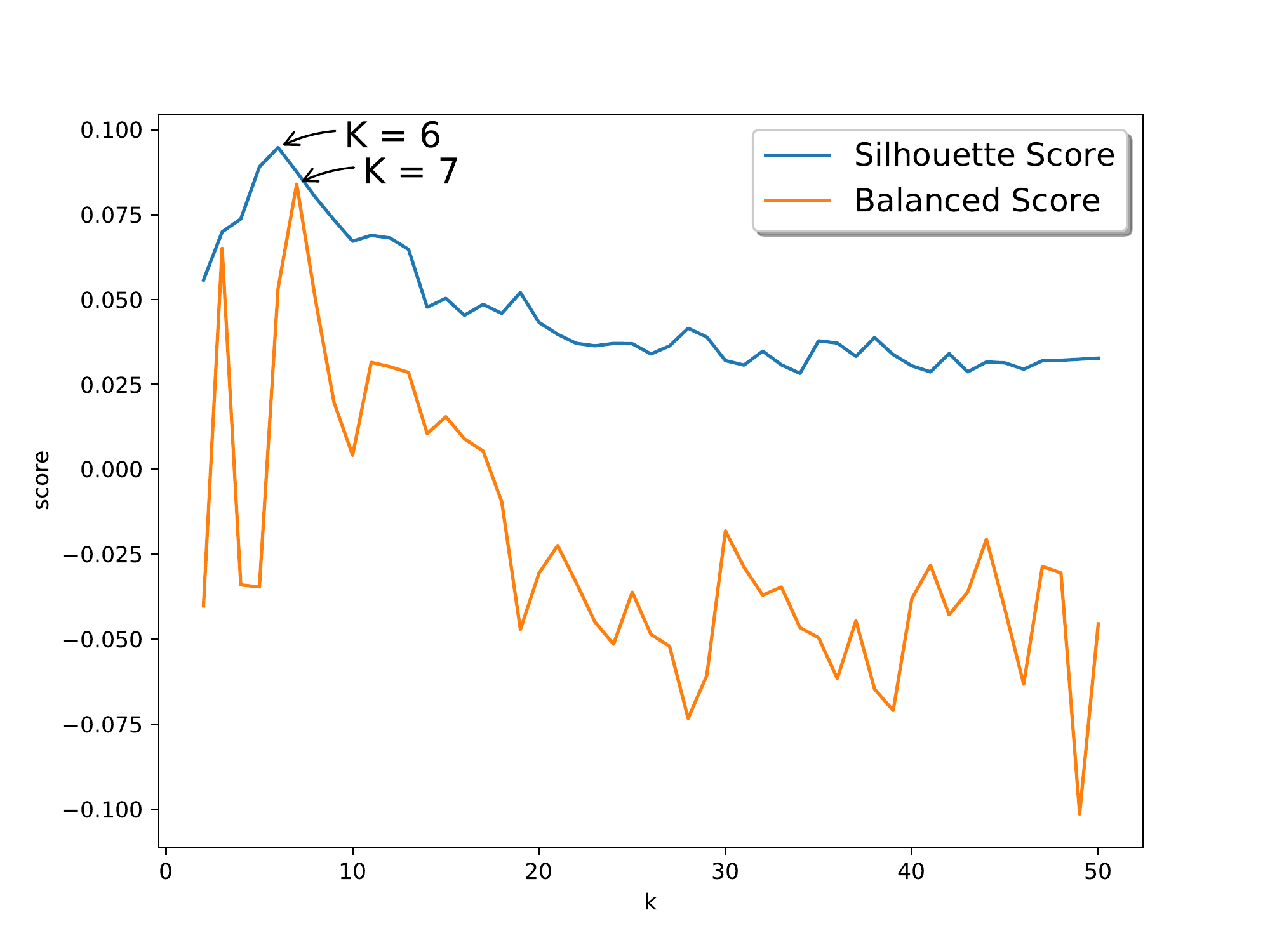}
  \caption{universal}
  \label{fig:use}
\end{subfigure}%
\caption{Changes of the Silhouette score and balanced score with different number of clusters using different representations.}
\label{fig:k-clusters}
\end{figure}

We adopt $K$-means for clustering and determine the number of clusters based on the metric of \textit{balanced score}.
As compared in Figure~\ref{fig:k-clusters}, using the balanced score leads to the correct number of clusters, (i.e., \textbf{7}) for either the \texttt{paraphrase} or \texttt{universal} representation, while the Silhouette score gives a correct $K$ only for the \texttt{paraphrase} representation.
Besides, the Silhouette score tends to have a level curve with slight changes under different $K$, while the balanced score changes dramatically since it gives higher penalty if a $K$ value leads to more unbalanced clusters.
We set $\lambda$ as 0.5 for computing the balanced scores in our experiments.


For qualitative comparison, we visualize the four types of semantic representations using t-SNE\cite{van2008visualizing}, where different colors represent different clusters. 
It can be seen that the \texttt{paraphrase} representation has better cluster separations than both \textit{nli-bert} and \textit{stsb-bert}, while the clusters using the \texttt{universal} representation show the best separations.
\begin{figure*}[htb]
\centering
\begin{subfigure}{0.239\linewidth}
  \centering
  \includegraphics[width=\linewidth]{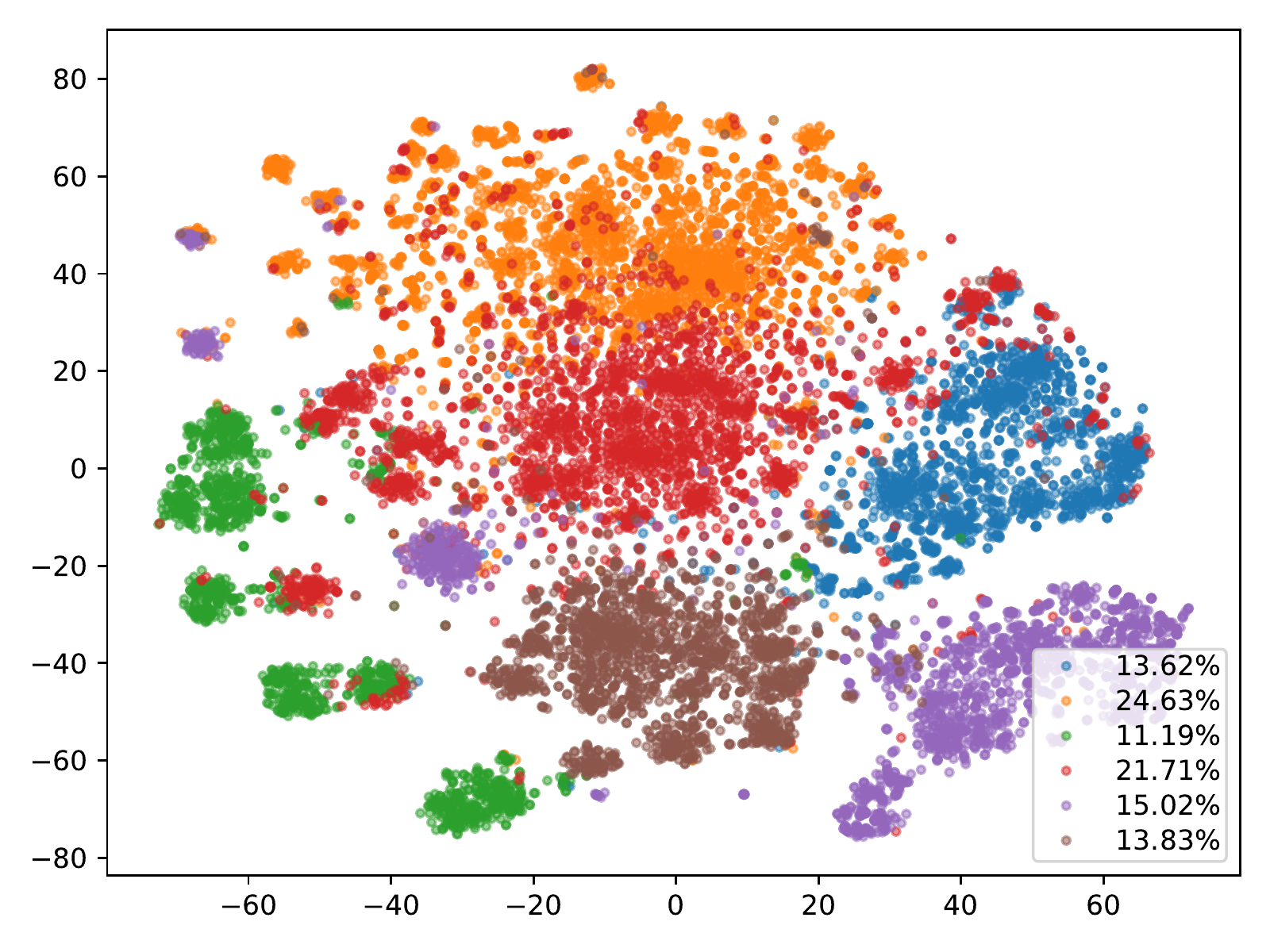}
  \caption{nli-bert}
  \label{fig:sfig2}
\end{subfigure}
\begin{subfigure}{0.239\linewidth}
  \centering
  \includegraphics[width=\linewidth]{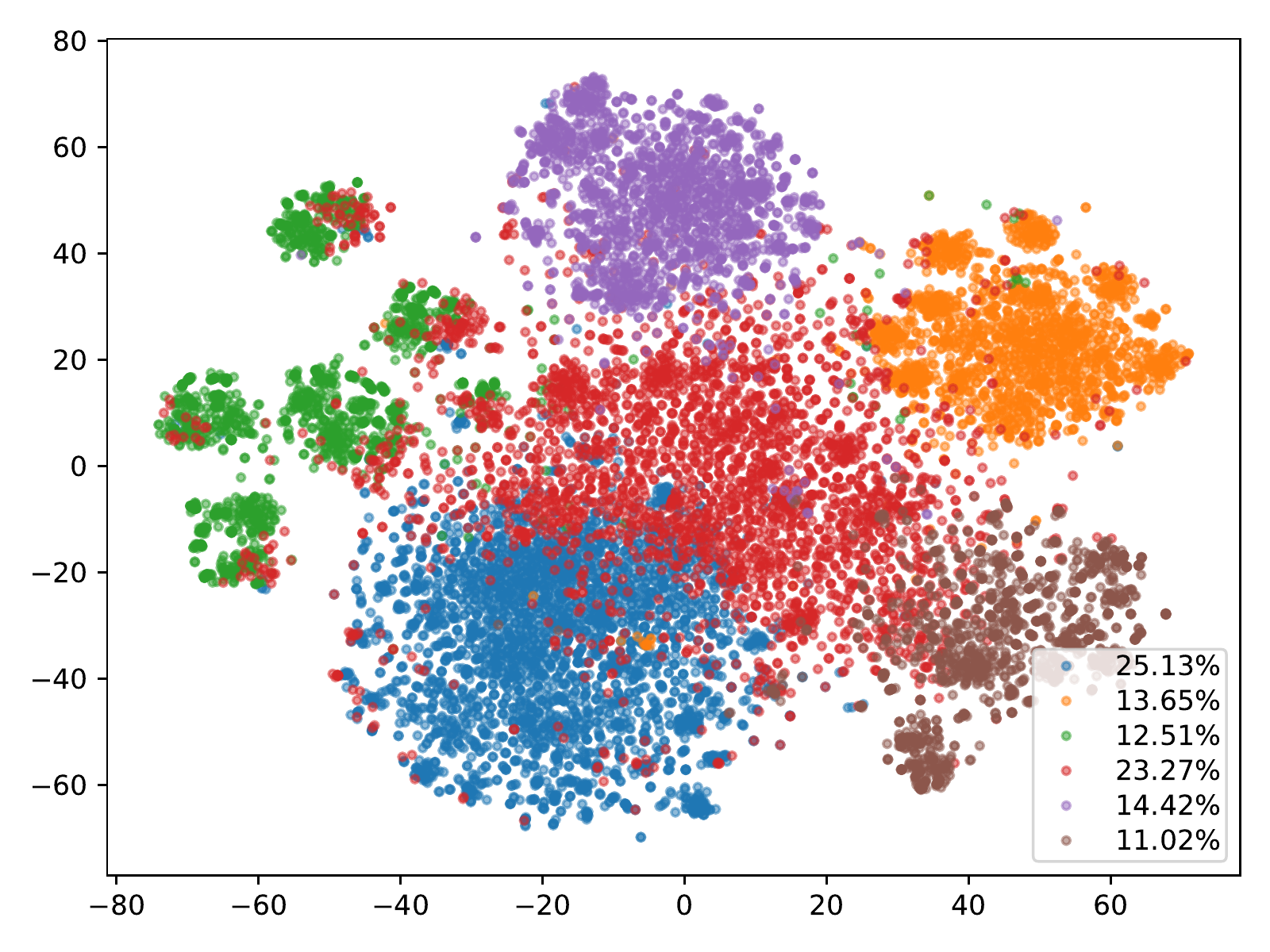}
  \caption{stsb-bert}
  \label{fig:sfig3}
\end{subfigure}%
\begin{subfigure}{0.239\linewidth}
  \centering
  \includegraphics[width=\linewidth]{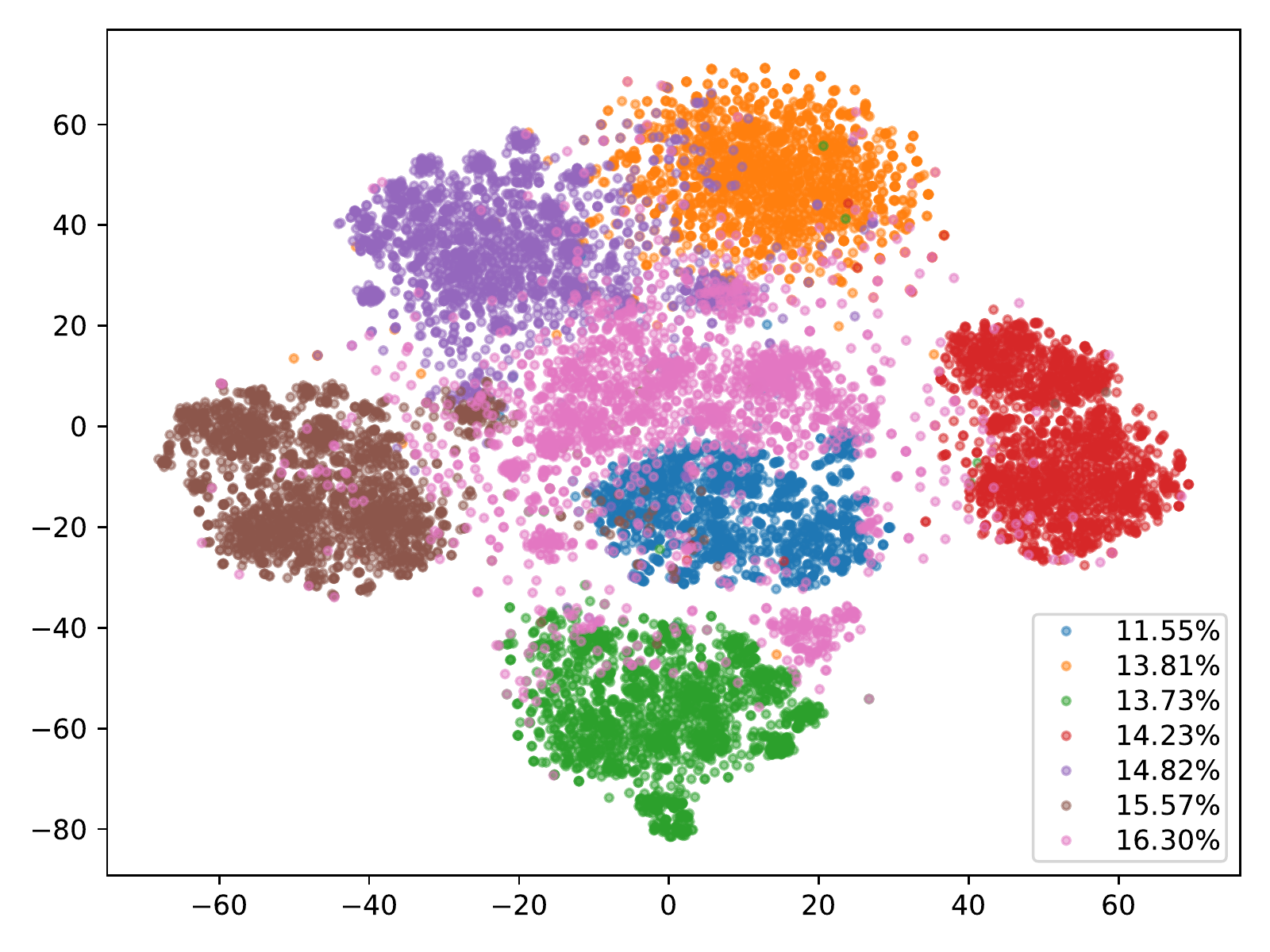}
  \caption{paraphrase}
  \label{fig:sfig4}
\end{subfigure}
\begin{subfigure}{0.239\linewidth}
  \centering
  \includegraphics[width=\linewidth]{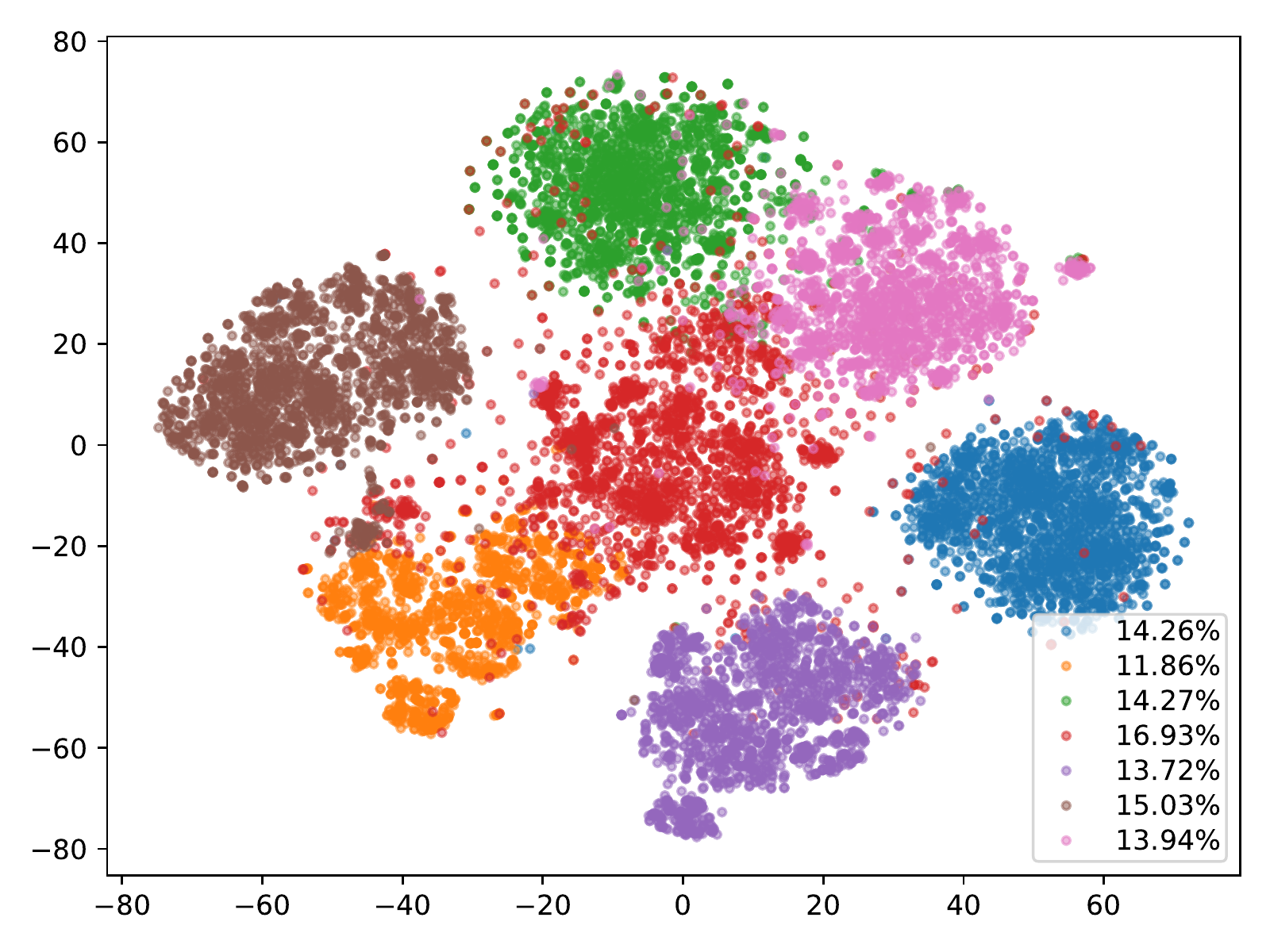}
  \caption{universal}
  \label{fig:tsne-use}
\end{subfigure}%
\vspace{-1em}
\caption{t-SNE visualization of clusters using different representations. The utterance percentage of each cluster is also shown.}
\label{fig:clusters}
\end{figure*}

\begin{table*}[htb]
\centering
\caption{Results of intent label generation by a dependency parser from the clusters obtained using the \textit{universal} representation.}
\label{tab:label-generation}
\resizebox{\textwidth}{!}{%
\begin{tabular}{clrlrr}
\hline
\textbf{Cluster} & \multicolumn{1}{c}{\textbf{Top 3 ACTION-OBJECT pairs}} & \multicolumn{1}{c}{\textbf{Percentage}} & \multicolumn{1}{c}{\textbf{Example Utterance}} & \multicolumn{1}{r}{\textbf{Generated}} & \multicolumn{1}{r}{\textbf{Ground-truth}} \\ \hline
0 & ('book-restaurant', 313), ('book-table', 281), ('book-spot', 131) & 36.90\% & i want to \textbf{book} a \textbf{restaurant} for nine with wifi & book-restaurant & BookRestaurant \\
1 & ('find-schedule', 164), ('be-schedule', 100), ('show-schedule', 98) & 22.14\% & \textbf{find} the movie \textbf{schedule} for close by films & find-schedule & SearchScreeningEvent \\
2 & ('add-tune', 160), ('add-song', 150), ('add-track', 134) & 22.57\% & \textbf{add} \textbf{tune} to sxsw fresh playlist & add-tune & AddToPlaylist \\
3 & ('find-show', 82), ('find-game', 50), ('show-creativity', 42) & 7.46\% & can you \textbf{find} me the spectres television \textbf{show} ? & find-show & SearchCreativeWork \\
4 & ('give-star', 133), ('rate-book', 104), ('rate-novel', 82) & 16.87\% & i \textbf{give} 4 of 6 \textbf{stars} for the saga severe mercy & give-star & RateBook \\
5 & ('be-weather', 395), ('be-forecast', 330), ('tell-forecast', 54) & 37.60\% & what \textbf{is} the \textbf{weather} for my current place & be-weather & GetWeather \\
6 & ('play-music', 424), ('play-song', 198), ('play-track', 75) & 36.28\% & please \textbf{play} some bill evans \textbf{music} & play-music & PlayMusic \\ \hline
\end{tabular}%
}
\end{table*}

\subsubsection{Stage II: Intent Label Generation}

Based on the clustering results from Stage I, we adopt a dependency parser to extract the ACTION-OBJECT pair for each utterance within each cluster.
Table~\ref{tab:label-generation} shows the top 3 ACTION-OBJECT pairs and the number of utterances having these pairs within each cluster, where the clustering results are obtained using the best \texttt{universal} representation. The third column presents the percentage of utterances covering the top 3 ACTION-OBJECT pairs. A higher percentage means the pair is more representative of the cluster.
Referring to each example utterance, it is observable to see that these pairs are meaningful labels for the clusters and match well with the ground-truth labels.
Note that although some datasets may not exhibit such ACTION-OBJECT patterns for intent labels, e.g., the very short informal conversations, the first stage of the framework is still applicable for identifying the unknown intents from the dataset and estimate the number of possible intents which can be very large in some complex datasets.

\subsubsection{Quantitative Evaluation}
Since the SNIPS dataset has ground-truth labels, we evaluate the proposed two-stage framework quantitatively by comparing the performance among the semantic representations and presenting the intent-level performance based on the best \texttt{universal} representation.

\noindent\textbf{Representation Comparisons.}
Recall that the representations are pre-trained on various datasets using different training objectives, hence, they may give different performance in the semantic clustering stage.
Table~\ref{tab:performance} shows the performance comparisons among different representations on the SNIPS dataset. It is clear that the pre-trained \texttt{universal} representation obtains the best performance on all the metrics.
This is also consistent with the cluster visualization results shown in Figure~\ref{fig:clusters}.

\begin{table}[htb]
\centering
\caption{Performance comparisons among representations.}
\label{tab:performance}
\resizebox{0.9\linewidth}{!}{%
\begin{tabular}{r|ccccc}
\hline
\textbf{Representation} & \multicolumn{1}{c}{\textbf{P}} & \multicolumn{1}{c}{\textbf{R}} & \multicolumn{1}{c}{\textbf{$F_1$}} & \multicolumn{1}{c}{\textbf{NMI}} & \multicolumn{1}{c}{\textbf{ARI}} \\ \hline
nli-bert & 0.682 & 0.635 & 0.647 & 0.597 & 0.505 \\
stsb-bert & 0.725 & 0.682 & 0.688 & 0.665 & 0.565 \\
paraphrase & 0.923 & 0.928 & 0.924 & 0.835 & 0.831 \\
universal & \textbf{0.934} & \textbf{0.940} & \textbf{0.935} & \textbf{0.865} & \textbf{0.855} \\ \hline
\end{tabular}%
}
\end{table}

\noindent\textbf{Intent-level Performance.}
We report the intent-level performance on the SNIPS dataset based on the \texttt{universal} representation.
As shown in Table~\ref{tab:snips}, all the intents have been successfully discovered and each ground-truth intent is semantically mapped to the corresponding generated intent label. Moreover, good performance is achieved for all the intents, since the universal representation can separate the utterances clearly.
We attribute the good clustering performance to the pre-trained models for obtaining semantic representations, and the metric of \textit{balanced score} in determining the number of clusters.

\begin{table}[htb]
\centering
\caption{Intent-level performance on the SNIPS dataset using the \texttt{universal} representation.}
\label{tab:snips}
\resizebox{0.9\linewidth}{!}{%
\begin{tabular}{r|r|ccc}
\hline
\textbf{Ground-truth} & \textbf{Generated} & \textbf{P} & \textbf{R} & \textbf{$F_1$} \\ \hline
AddToPlaylist & add-tune & 0.954 & 0.942 & 0.948 \\
BookRestaurant & book-restaurant & 0.987 & 0.991 & 0.989 \\
GetWeather & be-weather & 0.989 & 0.954 & 0.971 \\
PlayMusic & play-music & 0.894 & 0.931 & 0.912 \\
RateBook & give-star & 0.965 & 0.998 & 0.982 \\
SearchCreativeWork & find-show & 0.924 & 0.774 & 0.843 \\
SearchScreeningEvent & find-schedule & 0.826 & 0.990 & 0.901 \\ \hline
\multicolumn{2}{r|}{Average} & 0.934 & 0.940 & 0.935 \\ \hline
\end{tabular}%
}
\end{table}



\section{Conclusion}

The paper presents an unsupervised data-driven framework based on the clustering strategy in the semantic space to solve the open intent discovery problem. We first represent each utterance in the semantic space using the pre-trained Sentence-BERT model. Then, we apply a modified version of $K$-means clustering to all the utterances in the semantic space, where we introduce the balanced score which extends the Silhouette score by adding a penalty term to encourage balance among the intent clusters. As a result, the framework simultaneously accomplishes two tasks: (1) discover the intents from an unlabeled goal-oriented dialog corpus based on the clusters and (2) recognize the corresponding intent for each utterance from the corpus through its membership in the intent cluster. We further design a rule-based method to automatically generate an meaningful label in the form of ACTION-OBJECT for each discovered intent based on the dependency parsing on the utterances within the intent cluster. This allows a completely unsupervised intent discovery which can greatly save the costs of manual labeling for developing the intent understanding module in a dialog system. We evaluated the proposed framework with the SNIPS dataset and found that the discovered intents match exactly with the manually labeled intents. Moreover, the intents of the utterances are recognized in high precision and recall, showing that the proposed framework is effective in intent discovery.

For future work, it is worthwhile to extend the framework with other clustering methods for supporting imbalanced datasets, adopting new semantic representations and developing generation-based methods to generate meaningful cluster labels automatically.

\bibliographystyle{IEEEtran}
\bibliography{references}

\begin{thebibliography}{10}
\providecommand{\url}[1]{#1}
\csname url@samestyle\endcsname
\providecommand{\newblock}{\relax}
\providecommand{\bibinfo}[2]{#2}
\providecommand{\BIBentrySTDinterwordspacing}{\spaceskip=0pt\relax}
\providecommand{\BIBentryALTinterwordstretchfactor}{4}
\providecommand{\BIBentryALTinterwordspacing}{\spaceskip=\fontdimen2\font plus
\BIBentryALTinterwordstretchfactor\fontdimen3\font minus
  \fontdimen4\font\relax}
\providecommand{\BIBforeignlanguage}[2]{{%
\expandafter\ifx\csname l@#1\endcsname\relax
\typeout{** WARNING: IEEEtran.bst: No hyphenation pattern has been}%
\typeout{** loaded for the language `#1'. Using the pattern for}%
\typeout{** the default language instead.}%
\else
\language=\csname l@#1\endcsname
\fi
#2}}
\providecommand{\BIBdecl}{\relax}
\BIBdecl

\bibitem{perkins2019dialog}
H.~Perkins and Y.~Yang, ``Dialog intent induction with deep multi-view
  clustering,'' \emph{arXiv preprint arXiv:1908.11487}, 2019.

\bibitem{cai2017cnn}
R.~Cai, B.~Zhu, L.~Ji, T.~Hao, J.~Yan, and W.~Liu, ``An cnn-lstm attention
  approach to understanding user query intent from online health communities,''
  in \emph{2017 ieee international conference on data mining workshops
  (icdmw)}.\hskip 1em plus 0.5em minus 0.4em\relax IEEE, 2017, pp. 430--437.

\bibitem{vedula2019towards}
N.~Vedula, N.~Lipka, P.~Maneriker, and S.~Parthasarathy, ``Towards open intent
  discovery for conversational text,'' \emph{arXiv preprint arXiv:1904.08524},
  2019.

\bibitem{wang2015mining}
J.~Wang, G.~Cong, X.~Zhao, and X.~Li, ``Mining user intents in twitter: A
  semi-supervised approach to inferring intent categories for tweets,'' in
  \emph{Proceedings of the AAAI Conference on Artificial Intelligence},
  vol.~29, no.~1, 2015.

\bibitem{chatterjee2020intent}
A.~Chatterjee and S.~Sengupta, ``Intent mining from past conversations for
  conversational agent,'' in \emph{Proceedings of the 28th International
  Conference on Computational Linguistics}, 2020, pp. 4140--4152.

\bibitem{ashkan2009term}
A.~Ashkan and C.~L. Clarke, ``Term-based commercial intent analysis,'' in
  \emph{Proceedings of the 32nd international ACM SIGIR conference on Research
  and development in information retrieval}, 2009, pp. 800--801.

\bibitem{wang2013mining}
J.~Wang, W.~X. Zhao, H.~Wei, H.~Yan, and X.~Li, ``Mining new business
  opportunities: Identifying trend related products by leveraging commercial
  intents from microblogs,'' in \emph{Proceedings of the 2013 Conference on
  Empirical Methods in Natural Language Processing}, 2013, pp. 1337--1347.

\bibitem{pennington2014glove}
J.~Pennington, R.~Socher, and C.~D. Manning, ``Glove: Global vectors for word
  representation,'' in \emph{Proceedings of the 2014 conference on empirical
  methods in natural language processing (EMNLP)}, 2014, pp. 1532--1543.

\bibitem{peters2018deep}
M.~Peters, M.~Neumann, M.~Iyyer, M.~Gardner, C.~Clark, K.~Lee, and
  L.~Zettlemoyer, ``Deep contextualized word representations,'' in
  \emph{Proceedings of the 2018 Conference of the North American Chapter of the
  Association for Computational Linguistics: Human Language Technologies,
  Volume 1 (Long Papers)}, 2018, pp. 2227--2237.

\bibitem{devlin2018bert}
J.~Devlin, M.-W. Chang, K.~Lee, and K.~Toutanova, ``Bert: Pre-training of deep
  bidirectional transformers for language understanding,'' \emph{arXiv preprint
  arXiv:1810.04805}, 2018.

\bibitem{tur2011intent}
G.~Tur and L.~Deng, ``Intent determination and spoken utterance
  classification,'' \emph{Spoken Language Understanding: Systems for Extracting
  Semantic Information from Speech}, pp. 93--118, 2011.

\bibitem{liu2016attention}
B.~Liu and I.~Lane, ``Attention-based recurrent neural network models for joint
  intent detection and slot filling,'' \emph{arXiv preprint arXiv:1609.01454},
  2016.

\bibitem{meng2017dialogue}
L.~Meng and M.~Huang, ``Dialogue intent classification with long short-term
  memory networks,'' in \emph{National CCF Conference on Natural Language
  Processing and Chinese Computing}.\hskip 1em plus 0.5em minus 0.4em\relax
  Springer, 2017, pp. 42--50.

\bibitem{schuurmans2019intent}
J.~Schuurmans and F.~Frasincar, ``Intent classification for dialogue
  utterances,'' \emph{IEEE Intelligent Systems}, vol.~35, no.~1, pp. 82--88,
  2019.

\bibitem{khalil2019cross}
T.~Khalil, K.~Kie{\l}czewski, G.~C. Chouliaras, A.~Keldibek, and M.~Versteegh,
  ``Cross-lingual intent classification in a low resource industrial setting,''
  in \emph{Proceedings of the 2019 Conference on EMNLP-IJCNLP}, 2019, pp.
  6420--6425.

\bibitem{wang2020dialogue}
Y.~Wang, J.~Huang, T.~He, and X.~Tu, ``Dialogue intent classification with
  character-cnn-bgru networks,'' \emph{Multimedia Tools and Applications},
  vol.~79, no.~7, pp. 4553--4572, 2020.

\bibitem{firdaus2020deep}
M.~Firdaus, H.~Golchha, A.~Ekbal, and P.~Bhattacharyya, ``A deep multi-task
  model for dialogue act classification, intent detection and slot filling,''
  \emph{Cognitive Computation}, pp. 1--20, 2020.

\bibitem{zhang2016joint}
X.~Zhang and H.~Wang, ``A joint model of intent determination and slot filling
  for spoken language understanding.'' in \emph{IJCAI}, vol.~16, 2016, pp.
  2993--2999.

\bibitem{jansen2007determining}
B.~J. Jansen, D.~L. Booth, and A.~Spink, ``Determining the user intent of web
  search engine queries,'' in \emph{Proceedings of the 16th international
  conference on World Wide Web}, 2007, pp. 1149--1150.

\bibitem{cao2009context}
H.~Cao, D.~H. Hu, D.~Shen, D.~Jiang, J.-T. Sun, E.~Chen, and Q.~Yang,
  ``Context-aware query classification,'' in \emph{Proceedings of the 32nd
  international ACM SIGIR conference on Research and development in information
  retrieval}, 2009, pp. 3--10.

\bibitem{strohmaier2012acquiring}
M.~Strohmaier and M.~Kr{\"o}ll, ``Acquiring knowledge about human goals from
  search query logs,'' \emph{Information processing \& management}, vol.~48,
  no.~1, pp. 63--82, 2012.

\bibitem{goo2018slot}
C.-W. Goo, G.~Gao, Y.-K. Hsu, C.-L. Huo, T.-C. Chen, K.-W. Hsu, and Y.-N. Chen,
  ``Slot-gated modeling for joint slot filling and intent prediction,'' in
  \emph{Proceedings of the 2018 Conference of the North American Chapter of the
  Association for Computational Linguistics: Human Language Technologies,
  Volume 2 (Short Papers)}, 2018, pp. 753--757.

\bibitem{niu2019novel}
P.~Niu, Z.~Chen, M.~Song \emph{et~al.}, ``A novel bi-directional interrelated
  model for joint intent detection and slot filling,'' \emph{arXiv preprint
  arXiv:1907.00390}, 2019.

\bibitem{mikolov2013distributed}
T.~Mikolov, I.~Sutskever, K.~Chen, G.~Corrado, and J.~Dean, ``Distributed
  representations of words and phrases and their compositionality,''
  \emph{arXiv preprint arXiv:1310.4546}, 2013.

\bibitem{liu2019roberta}
Y.~Liu, M.~Ott, N.~Goyal, J.~Du, M.~Joshi, D.~Chen, O.~Levy, M.~Lewis,
  L.~Zettlemoyer, and V.~Stoyanov, ``Roberta: A robustly optimized bert
  pretraining approach,'' \emph{arXiv preprint arXiv:1907.11692}, 2019.

\bibitem{reimers2019sentence}
N.~Reimers, I.~Gurevych, N.~Reimers, I.~Gurevych, N.~Thakur, N.~Reimers,
  J.~Daxenberger, I.~Gurevych, N.~Reimers, I.~Gurevych \emph{et~al.},
  ``Sentence-bert: Sentence embeddings using siamese bert-networks,'' in
  \emph{Proceedings of the 2019 Conference on Empirical Methods in Natural
  Language Processing}.\hskip 1em plus 0.5em minus 0.4em\relax Association for
  Computational Linguistics, 2019.

\bibitem{macqueen1967some}
J.~MacQueen \emph{et~al.}, ``Some methods for classification and analysis of
  multivariate observations,'' in \emph{Proceedings of the fifth Berkeley
  symposium on mathematical statistics and probability}, vol.~1, no.~14.\hskip
  1em plus 0.5em minus 0.4em\relax Oakland, CA, USA, 1967, pp. 281--297.

\bibitem{reynolds2009gaussian}
D.~A. Reynolds, ``Gaussian mixture models.'' \emph{Encyclopedia of biometrics},
  vol. 741, pp. 659--663, 2009.

\bibitem{ester1996density}
M.~Ester, H.-P. Kriegel, J.~Sander, X.~Xu \emph{et~al.}, ``A density-based
  algorithm for discovering clusters in large spatial databases with noise.''
  in \emph{Kdd}, vol.~96, no.~34, 1996, pp. 226--231.

\bibitem{rousseeuw1987silhouettes}
P.~J. Rousseeuw, ``Silhouettes: a graphical aid to the interpretation and
  validation of cluster analysis,'' \emph{Journal of computational and applied
  mathematics}, vol.~20, pp. 53--65, 1987.

\bibitem{chen2016zero}
Y.-N. Chen, D.~Hakkani-T{\"u}r, and X.~He, ``Zero-shot learning of intent
  embeddings for expansion by convolutional deep structured semantic models,''
  in \emph{2016 IEEE International Conference on Acoustics, Speech and Signal
  Processing (ICASSP)}.\hskip 1em plus 0.5em minus 0.4em\relax IEEE, 2016, pp.
  6045--6049.

\bibitem{xia2018zero}
C.~Xia, C.~Zhang, X.~Yan, Y.~Chang, and S.~Y. Philip, ``Zero-shot user intent
  detection via capsule neural networks,'' in \emph{Proceedings of the 2018
  Conference on Empirical Methods in Natural Language Processing}, 2018, pp.
  3090--3099.

\bibitem{liu2019reconstructing}
H.~Liu, X.~Zhang, L.~Fan, X.~Fu, Q.~Li, X.-M. Wu, and A.~Y. Lam,
  ``Reconstructing capsule networks for zero-shot intent classification,'' in
  \emph{Proceedings of the 2019 Conference on Empirical Methods in Natural
  Language Processing and the 9th International Joint Conference on Natural
  Language Processing (EMNLP-IJCNLP)}, 2019, pp. 4799--4809.

\bibitem{chen2013identifying}
Z.~Chen, B.~Liu, M.~Hsu, M.~Castellanos, and R.~Ghosh, ``Identifying intention
  posts in discussion forums,'' in \emph{Proceedings of the 2013 conference of
  the North American chapter of the association for computational linguistics:
  human language technologies}, 2013, pp. 1041--1050.

\bibitem{hamerly2004learning}
G.~Hamerly and C.~Elkan, ``Learning the k in k-means,'' \emph{Advances in
  neural information processing systems}, vol.~16, pp. 281--288, 2004.

\bibitem{everitt2002cambridge}
B.~Everitt and A.~Skrondal, \emph{The Cambridge dictionary of
  statistics}.\hskip 1em plus 0.5em minus 0.4em\relax Cambridge University
  Press Cambridge, 2002, vol. 106.

\bibitem{bouman1997cluster}
C.~BOUMAN, ``Cluster: An unsupervised algorithm for modeling gaussian
  mixtures,'' \emph{http://www. ece. purdue. edu/\~{} bouman}, 1997.

\bibitem{honnibal2015improved}
M.~Honnibal and M.~Johnson, ``An improved non-monotonic transition system for
  dependency parsing,'' in \emph{Proceedings of the 2015 conference on
  empirical methods in natural language processing}, 2015, pp. 1373--1378.

\bibitem{coucke2018snips}
A.~Coucke, A.~Saade, A.~Ball, T.~Bluche, A.~Caulier, D.~Leroy, C.~Doumouro,
  T.~Gisselbrecht, F.~Caltagirone, T.~Lavril \emph{et~al.}, ``Snips voice
  platform: an embedded spoken language understanding system for
  private-by-design voice interfaces,'' \emph{arXiv preprint arXiv:1805.10190},
  2018.

\bibitem{cer2018universal}
D.~Cer, Y.~Yang, S.-y. Kong, N.~Hua, N.~Limtiaco, R.~S. John, N.~Constant,
  M.~Guajardo-C{\'e}spedes, S.~Yuan, C.~Tar \emph{et~al.}, ``Universal sentence
  encoder,'' \emph{arXiv preprint arXiv:1803.11175}, 2018.

\bibitem{knops2006normalized}
Z.~F. Knops, J.~A. Maintz, M.~A. Viergever, and J.~P. Pluim, ``Normalized
  mutual information based registration using k-means clustering and shading
  correction,'' \emph{Medical image analysis}, vol.~10, no.~3, pp. 432--439,
  2006.

\bibitem{rand1971objective}
W.~M. Rand, ``Objective criteria for the evaluation of clustering methods,''
  \emph{Journal of the American Statistical association}, vol.~66, no. 336, pp.
  846--850, 1971.

\bibitem{van2008visualizing}
L.~Van~der Maaten and G.~Hinton, ``Visualizing data using t-sne.''
  \emph{Journal of machine learning research}, vol.~9, no.~11, 2008.

\end{thebibliography}

\end{document}